\documentclass{article}
\usepackage[utf8]{inputenc}

\usepackage{amsmath}
\usepackage{mathcomp}
\usepackage{graphicx}
\usepackage{url}
\graphicspath{ {./figures/} }
\usepackage{biblatex}
\title{Efficient Computation of Higher Order 2D Image Moments using the Discrete Radon Transform}
\author{William Diggin
     \and 
     Michael Diggin}
\date{September 2020}

\begin{document}

\maketitle

\begin{abstract}
Geometric moments and moment invariants of image artifacts have many uses in computer vision applications, e.g. shape classification or object position and orientation. Higher order moments are of interest to provide additional feature descriptors, to measure kurtosis or to resolve n-fold symmetry. This paper provides the method and practical application to extend an efficient algorithm, based on the Discrete Radon Transform, to generate moments greater than the $3^{rd}$ order. The mathematical fundamentals are presented, followed by relevant implementation details. Results of scaling the algorithm based on image area and its computational comparison with a standard method demonstrate the efficacy of the approach. 
\end{abstract}

\section{Introduction}
In his seminal paper, Hu \cite{HuArticle} defined 10 geometric moments up to the $3^{rd}$ order which are used to derive 7 moment invariants, e.g. invariant to position, scale, orientation, etc. These invariants are useful as feature descriptors for objects in images. The moment generating equation for a 2-D image $I(i,j)$ of order $(p+q)$ is: 
\begin{equation}
\label{base_moments}
    M_{pq} = \sum_{i, j} I(i,j) i^p j^q 
\end{equation}

Equivalent $4^{th}$ order moments have been utilised by Prokop et al \cite{prokop1992survey} to measure kurtosis. Higher order moments have been shown to be useful discriminators for n-fold rotational symmetry \cite{flusser}. Some previous work in image moments focuses on the algorithms and the computational cost of those algorithms to generate moments up to the $3^{rd}$ order; the goal to optimise execution time with little or no loss to accuracy. An algorithm based on the Discrete Radon Transform \cite{DRT} exhibits reduced complexity and faster execution time in comparison to several other techniques, while retaining accuracy \cite{diggin2020using}. Other early work in using the Discrete Radon Transform for image moments can be attributed to Gindi et al \cite{gindi1984optical} and Shen et al \cite{shen1996fast}.

In section \ref{drtalgorithm}, we recall the Discrete Radon Transform and its application to image moments calculation. In section \ref{edrtalgorithm}, we outline the approach to extend the DRT algorithm to the $4^{th}$ order. Then, some practical aspects are disclosed when implementing the algorithm, followed by a presentation of comparative results that demonstrate the efficacy of the algorithm relative to the OpenCV Library \cite{opencv_library}. Finally, in section \ref{analysis}, we conduct some analysis of the results and capture some of the learnings from the approach.

\section{Discrete Radon Transform Algorithm} \label{drtalgorithm}
The Radon Transform is the set of projections of a function $f(x,y)$ onto line integrals at various angles over a range, $0\leq\theta<\pi$:
\[
    \mathcal{R}(x^{'},\theta)=\int f(x,y)dy^{'} ~~~where \begin{pmatrix} x^{'}\\ y^{'}\end{pmatrix}=\begin{pmatrix}
    \cos\theta &\sin\theta\\ \!-\!\sin\theta &\cos\theta \end{pmatrix}
    \begin{pmatrix} x\\ y\end{pmatrix}
\]

For the Discrete Radon Transform (DRT), let $\mathcal{R}^{x:y}$ be a projection at angle $\theta=\tan^{-1}\frac{y}{x}$ of an image $I(i,j)$, whose size is $M\!\cdot\! N$.
The quantities $x\!:\!y$ are the ratio of horizontal to vertical units, equivalent to slope. For an image, these units are pixels. Given this, we can efficiently compute the projections of the image at angles 0\textdegree, 90\textdegree, 45\textdegree (diagonal) and 135\textdegree (anti-diagonal) as follows: 
\begin{align*}
    \mathcal{R}^{1:0}[k]&=\sum_{j} I(k,j) ~~~where\  k\!=\!0...(M\!-\!1)\\
    \mathcal{R}^{0:1}[k]&=\sum_{j} I(j,k) ~~~where\  k\!=\!0...(N\!-\!1)\\
    \mathcal{R}^{1:1}[k]&=\sum_{i+j=k} I(i,j) ~~~where\  k\!=\!0...(M\!+\!N\!-\!1)\\
    \mathcal{R}^{-1:1}[k]&=\sum_{j-i=k} I(i,j) ~~~where\  k\!=\!(\!-\!N\!+\!1)...0...(M\!-\!1)
\end{align*}
Note negative indexing for $\mathcal{R}^{-1:1}$.

Let $M_r^{x:y}$ be the moment, of order $r$, for a 1-D projection of a DRT whose pixel ratio is $x\!:\!y$. Then the equivalent 1-D moments over the projection are given by:
\begin{equation}
    \label{Mxy}
    M_r^{x:y}=\sum_{k}\mathcal{R}^{x:y}[k]\cdot k^r
\end{equation}

It was shown that various 1-D moments over the 4 projections $\mathcal{R}^{1:0}$, $\mathcal{R}^{0:1}$, $\mathcal{R}^{1:1}$ and $\mathcal{R}^{-1:1}$, can be used to compute the 10 Hu 2-D moments up to $3^{rd}$ order \cite{diggin2020using}. The stated benefit is a reduction in computation time since the complexity of the DRT algorithm is 
$\mathcal{O}(n)$ versus $\mathcal{O}(n^{2})$ for the 2-D computational approach.

\section{Extended DRT Algorithm} \label{edrtalgorithm}
Extending the DRT algorithm to generate \emph{some} $4^{th}$ order 2-D moments does not require additional projections. For example, the moments, $M_{40}$ and $M_{04}$, are trivial to compute from \eqref{Mxy} using the horizontal and vertical projections as follows:
\begin{equation}
    \label{M40}
    M_{40}=M_{4}^{1:0}
\end{equation}
\begin{equation}
    \label{M04}
    M_{04}=M_{4}^{0:1}
\end{equation}

The $4^{th}$ order $M_{22}$ moment can be computed from the $4^{th}$ order diagonal and anti-diagonal 1-D moments as follows. Take the diagonal:
\begin{align}
\nonumber
    M_4^{1:1}&=\sum_{k}\mathcal{R}^{1:1}[k]\cdot k^4\\
\nonumber
    M_4^{1:1}&=\sum_{k}\sum_{i+j=k}I(i,j)\cdot k^4\\
\nonumber
    M_4^{1:1}&=\sum_{i,j}I(i,j)(i+j)^4\\
\nonumber
    M_4^{1:1}&=\sum_{i,j}I(i,j)(i^4+j^4+4i^3j+4ij^3+6i^2j^2)\\
    \label{M4_11}
    M_4^{1:1}&=M_{40}+M_{04}+4M_{31}+4M_{13}+6M_{22}
\end{align}
Similarly, for the anti-diagonal:
\begin{align}
\nonumber
    M_4^{-1:1}&=\sum_{k}\sum_{j-i=k}I(i,j)\cdot k^4\\
    \label{M4_-11}
    M_4^{-1:1}&=M_{40}+M_{04}-4M_{31}-4M_{13}+6M_{22}
\end{align}
Adding \eqref{M4_11} and \eqref{M4_-11} and rearranging:
\begin{equation}
    \label{M22}
    M_{22}=(M_4^{1:1}+M_4^{-1:1}-2M_{40}-2M_{04})/12
\end{equation}

\begin{figure}
\centering
\includegraphics[scale=1]{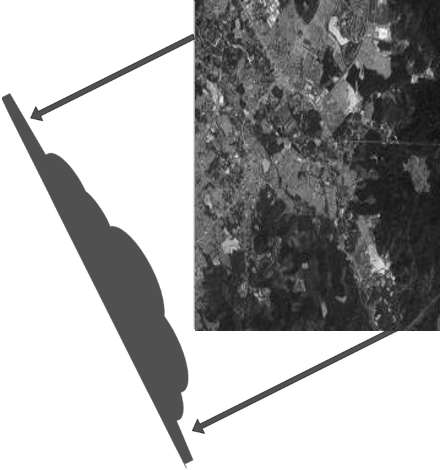}
\caption{A Projection of the Radon Transform for Slope=2.}
\text{(projection at angle $\approx$63.4\textdegree)}
\label{fig:combinedslope2}
\end{figure}

Computing $M_{31}$ and $M_{13}$ does require an additional projection. A convenient projection is:
\begin{equation}
    \label{R12}
    \mathcal{R}^{1:2}[k]=\sum_{i+2j=k} I(i,j) ~~~where\
    k\!=\!0...(M\!+\!2N\!-\!1)
\end{equation}
This is equivalent to a projection at a slope of 2 (angle $\approx$63.4\textdegree) onto a line integral, as depicted in figure \ref{fig:combinedslope2}. Then, to derive $M_{31}$ and $M_{13}$, we use \eqref{Mxy} and \eqref{R12} and note:
\begin{align}
\nonumber
    M_4^{1:2}&=\sum_{k}\mathcal{R}^{1:2}[k]\cdot k^4\\
\nonumber
    M_4^{1:2}&=\sum_{k}\sum_{i+2j=k}I(i,j)\cdot k^4\\
\nonumber
    M_4^{1:2}&=\sum_{i,j}I(i,j)(i+2j)^4\\
\nonumber
    M_4^{1:2}&=\sum_{i,j}I(i,j)(i^4+16j^4+8i^3j+32ij^3+24i^2j^2)\\
    \label{M4_12}
    M_4^{1:2}&=M_{40}+16M_{04}+8M_{31}+32M_{13}+24M_{22}
\end{align}
Combining \eqref{M4_11} with \eqref{M4_12} and rearranging:
\begin{equation}
    \label{M13}
    M_{13}=(M_4^{1:2}-2M_4^{1:1}+M_{40}-14M_{04}-12M_{22})/24
\end{equation}
By further substitution into \eqref{M4_12}:
\begin{equation}
    \label{M31}
    M_{31}=(M_4^{1:2}-M_{40}-16M_{04}-32M_{13}-24M_{22})/8
\end{equation}
\section{Implementation Notes} \label{implementation}
\begin{figure}
\centering
\includegraphics[scale=0.5]{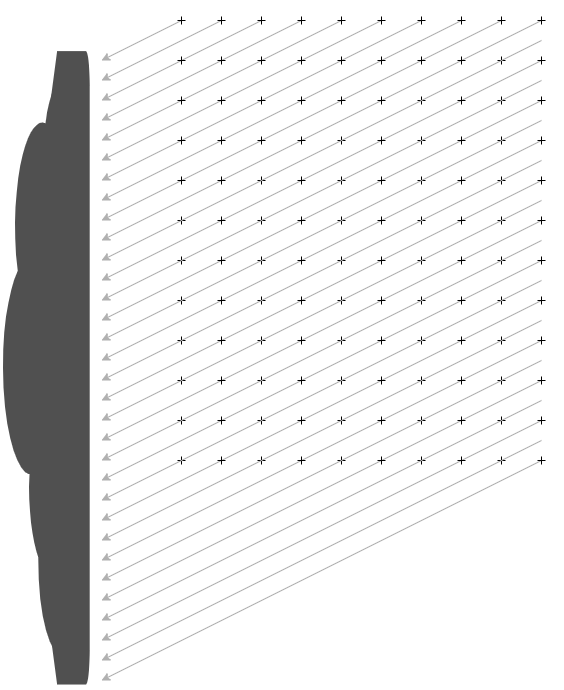}
\caption{A Projection of the \underline{Discrete} Radon Transform of Image of Size $M\cdot N$.}
\text{(for Slope=2, projection length = $M\!+\!2N$)}
\label{fig:discreteslope2}
\end{figure}

In practical terms, the projection at slope 2 can be discrete (see figure \ref{fig:discreteslope2}) resulting in a linear projection. We benefit from not requiring interpolation and the associated cost of that computation. This is at the expense of memory, as the resulting projection is of length $M\!+\!2N$.

When executing the DRT algorithm, the underlying processor capabilities are a vital component in efficient computation. Over the years, the cost of integer multiplication has improved to equate to the cost of integer addition – in line with advancement in processor technologies and their architectures \cite{wiki:processors}. So too has floating-point arithmetic. On the Intel processors, the x64 architecture, SIMD (Single Instruction, Multiple Data) and further instruction sets (SSE – Streaming SIMD Extensions or AVX – Advanced Vector Extensions) provide optimisations where computations can be parallelised or pipelined, resulting in greater algorithm performance. In the example presented herein, we encode the algorithm to rely on and take advantage of the AVX optimisations provided through compiler settings (e.g. \emph{/arch:AVX2} in the Microsoft C++ compiler).

\section{Results} \label{results}
To demonstrate the improvement from the DRT algorithm, we compare it to an extended OpenCV implementation – where the base algorithm used in the OpenCV library is extended to compute the $4^{th}$ order moments. Using a large sample size and selecting the fastest exemplar, the comparison of execution times was gathered for differing image sizes. Both algorithms were executed on an Intel Core i5 $7^{th}$ Gen CPU. Although we are concerned mainly with relative times, execution with and without AVX processor optimisations were also noted.

\begin{figure}[ht]
\centering
\includegraphics[scale=0.65]{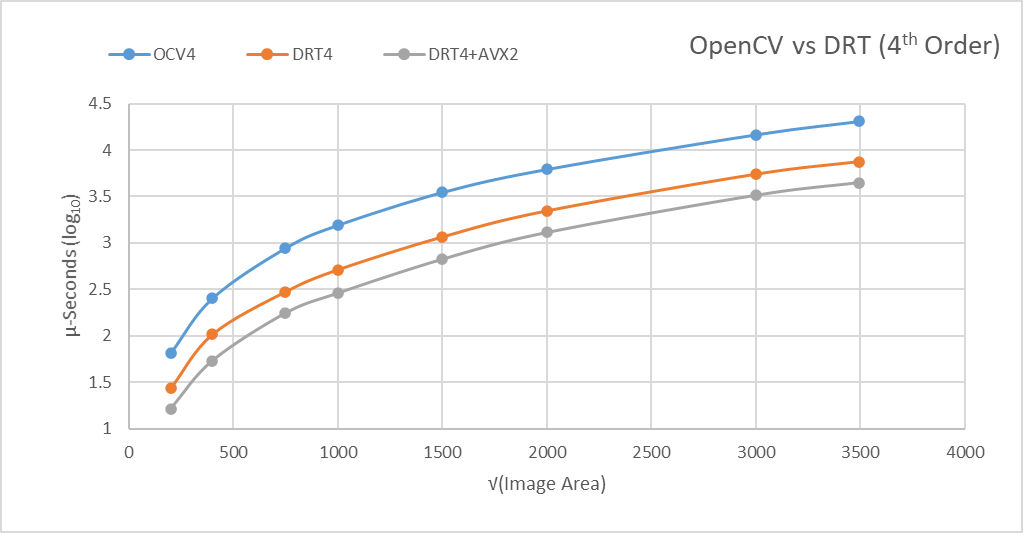}
\caption{Comparative Computational Cost Vs Image Area for OpenCV and DRT Algorithms.}
\label{fig:plot3}
\end{figure}

Table \ref{table:1} and figure \ref{fig:plot3} show the measurements and plot comparing the time needed to compute the raw image moments of a given image size. The smallest image used was a 200x200 pixel image, with the largest at 4032x3024 pixels. The y-axis illustrates the execution time in $\mu$-seconds (log base 10 for easier plotting). The x-axis is the square root of the number of pixels in the image (as this is a better method to illustrate the performance in terms of image size). We also include the DRT method with and without SSE2/AVX optimisation (provided through the compiler settings). 
\begin{table}[ht]
\begin{center}
\begin{tabular}{ |r| r| r| r| r| }
\hline
\multicolumn{2}{|r|}{\textbf{Image Size}} & \textbf{OCV4} & \textbf{DRT4} & \textbf{DRT4+AVX2}\\
\multicolumn{2}{|r|}{(pixels)} & $\mu$sec & $\mu$sec & $\mu$sec\\
\hline\hline
4032 & 3024 & 20363 & 7426 & 4455\\
\hline
3000 & 3000 & 14644 & 5444 & 3260\\
\hline
2000 & 2000 & 6209 & 2194 & 1292\\
\hline
1500 & 1500 & 3500 & 1151 & 665\\
\hline
1000 & 1000 & 1552 & 510 & 287\\
\hline
750 & 750 & 878 & 296 & 175\\
\hline
400 & 400 & 253 & 103 & 54\\
\hline
200 & 200 & 64 & 27 & 16\\
\hline
\end{tabular}
\caption{Comparative Computational Time vs Image Size for OpenCV and DRT Algorithms.}
\label{table:1}
\end{center}
\end{table}
\section{Analysis} \label{analysis}
The naïve implementation to compute all moments up to the $4^{th}$ order using equation \eqref{base_moments} requires $(15M\! \cdot \!N)$ multiplications and $(15M\! \cdot \!N)$ additions. From inspection of the OpenCV implementation, the same computation requires $(4M\! \cdot \!N\!+\!12N)$ multiplications with $(5M\! \cdot \!N\!+\!15N)$ additions. The DRT algorithm presented herein needs only $(10M\!+\!11N)$ multiplications and $(5M\! \cdot \!N\!+\!11M\!+\!11N)$ additions – overall a significant improvement. A 4x–5.5x performance gain relative to the OpenCV implementation is achieved across various image sizes.

To extend the approach to higher order moments, e.g. $5^{th}$ or $6^{th}$ order, requires further projections. The standard horizontal and vertical projections will continue to generate the desired $M_{i0}$, $M_{0i}$ moments. The diagonal and anti-diagonal projections will contribute to computing the $M_{ij}$ moments, but the number of unknowns will exceed the number of equations to resolve all moments, as exemplified for the $4^{th}$ order moment computation. In that case, ratios of 2:1, 1:2, -1:2 and 1:-2 are convenient inputs to the transform. While the $4^{th}$ order moments have 5 unknowns ($M_{40}$, $M_{04}$, $M_{22}$, $M_{31}$, $M_{13}$) to resolve and required 5 equations from 5 1-D moments ($M_4^{1:0}$, $M_4^{0:1}$, $M_4^{1:1}$, $M_4^{-1:1}$ and $M_4^{1:2}$), the $5^{th}$ order moment set requires 6 equations for 6 unknowns ($M_{50}$, $M_{05}$, $M_{41}$, $M_{14}$, $M_{32}$, $M_{23}$); the $6^{th}$ order requiring 7 equations for ($M_{60}$, $M_{06}$, $M_{51}$, $M_{15}$, $M_{42}$, $M_{24}$, $M_{33}$) and so on.

The selection of slope is not arbitrary. Indeed, it is selected to provide for a discrete summation into a linear vector. The desired slopes are derived from the integer ratio of vertical to horizontal pixels. For example, 3:1 or 3:2 and their corollary, 1:3 or 2:3, provide for convenient summations that do not require interpolation nor are so large as to make them unwieldy. Those ratio choices can be further expanded upon via reflections about the X- and Y-axes.

In execution, particularly on Intel CPUs, the cost of multiplication versus addition, while not significant, does contribute additional execution time. With optimisation and careful algorithm construction, we take advantage of pipelining and parallelism to counter that margin. Use of the AVX optimisations contributes a $\approx1.7x$ gain for the DRT algorithm.

Relative to the OpenCV algorithm, much of the computational savings arise from reducing the total number of operations, either multiplication or addition. The OpenCV algorithm has nearly twice as many operations as the DRT algorithm. The contributions from a reduction in the number of total operations, a reduction in the number of multiplications and careful algorithm/processor optimisations leads to a significant computational advantage.

\section{Conclusion} \label{conclusion}

The DRT algorithm provides increasing computational advantages as either of image size and/or moment order increases. Extending the algorithm to compute the full $4^{th}$ order moment set requires an additional projection in addition to those required to compute all the $3^{rd}$ order moments. The resulting computations are simple and elegant. The $4^{th}$ order moments of 5 projections generate all 15 2-D moments up to the $4^{th}$ order exactly. The DRT algorithm reduces the net multiplicative complexity from $\mathcal{O}(n^{2})$ to $\mathcal{O}(n)$ although the additive complexity remains $\mathcal{O}(n^{2})$. The approach allows a practical means to extend the algorithm to higher order moments, while retaining computational efficiency.

Further work in this area may seek to determine optimal relationships between higher order moments and the Inverse Radon Transform, e.g. an algorithm to reconstruct an image from its 1-D moments.

\printbibliography

\end{document}